\documentclass[10pt,twocolumn,letterpaper]{article}

\usepackage{iccv}
\usepackage{times}
\usepackage{epsfig}
\usepackage{graphicx}
\usepackage{amsmath}
\usepackage{amssymb}
\usepackage{xcolor}

\usepackage{cite}

\usepackage{booktabs}
\usepackage{tabularx}
\usepackage{multirow}

\newcolumntype{Y}{>{\centering\arraybackslash}X}
\newcolumntype{P}[1]{>{\centering\arraybackslash}p{#1}}
\newcolumntype{C}[1]{>{\centering\let\newline\\\arraybackslash\hspace{0pt}}m{#1}}

\DeclareMathOperator*{\argmin}{arg\,min}

\usepackage{pifont}%
\newcommand{\cmark}{\ding{51}}%
\newcommand{\xmark}{\ding{55}}%

\usepackage[pagebackref=true,breaklinks=true,letterpaper=true,colorlinks,bookmarks=false]{hyperref}

\iccvfinalcopy %

\def \paravspace {-0.7\baselineskip}

\ificcvfinal\fi

\begin{document}

\title{End-to-End Spatio-Temporal Action Localisation with Video Transformers}

\author{Alexey Gritsenko \quad Xuehan Xiong \quad Josip Djolonga \quad Mostafa Dehghani \\ \quad Chen Sun \quad Mario Lučić  \quad  Cordelia Schmid \quad Anurag Arnab \\
	Google Research \\
	{\tt\small \{agritsenko, aarnab\}@google.com}
}

\maketitle
\ificcvfinal\thispagestyle{empty}\fi

\begin{abstract}
The most performant spatio-temporal action localisation models use external person proposals and complex external memory banks.
We propose a fully end-to-end, purely-transformer based model that directly ingests an input video, and outputs tubelets -- a sequence of bounding boxes and the action classes at each frame.
Our flexible model can be trained with either sparse bounding-box supervision on individual frames, or full tubelet annotations. And in both cases, it predicts coherent tubelets as the output.
Moreover, our end-to-end model requires no additional pre-processing in the form of proposals, or post-processing in terms of non-maximal suppression.
We perform extensive ablation experiments, and significantly advance the state-of-the-art results on four different spatio-temporal action localisation benchmarks
with both sparse keyframes and full tubelet annotations.
\end{abstract}

\section{Introduction}

Spatio-temporal action localisation is an important problem with applications in advanced video search engines, robotics and security among others.
It is typically formulated in one of two ways: Firstly, predicting the bounding boxes and actions performed by an actor at a single keyframe given neighbouring frames as spatio-temporal context~\cite{gu_cvpr_2018,li2020ava}.
Or alternatively, predicting a sequence of bounding boxes and actions (\ie ``tubes''), for each actor at each frame in the video~\cite{soomro_arxiv_2012,jhuang2013towards}. %

The most performant models~\cite{pan2021actor,arnab2022beyond,wu2022memvit,feichtenhofer_iccv_2019}, particularly for the first, keyframe-based formulation of the problem, employ a two-stage pipeline inspired by the Fast-RCNN object detector~\cite{girshick_iccv_2015}:
They first run a separate person detector to obtain proposals. %
Features from these proposals are then aggregated and classified according to the actions of interest.
These models have also been supplemented with memory banks containing long-term contextual information from other frames~\cite{wu_cvpr_2019,wu2022memvit,pan2021actor,tang2020asynchronous}, and/or detections of other potentially relevant objects~\cite{tang2020asynchronous,arnab2021unified} to capture additional scene context, achieving state-of-the-art results.

\begin{figure}[t]
	\vspace{-0.5\baselineskip}
	\includegraphics[width=\linewidth, trim={0 7.9cm 18.4cm 0}, clip]{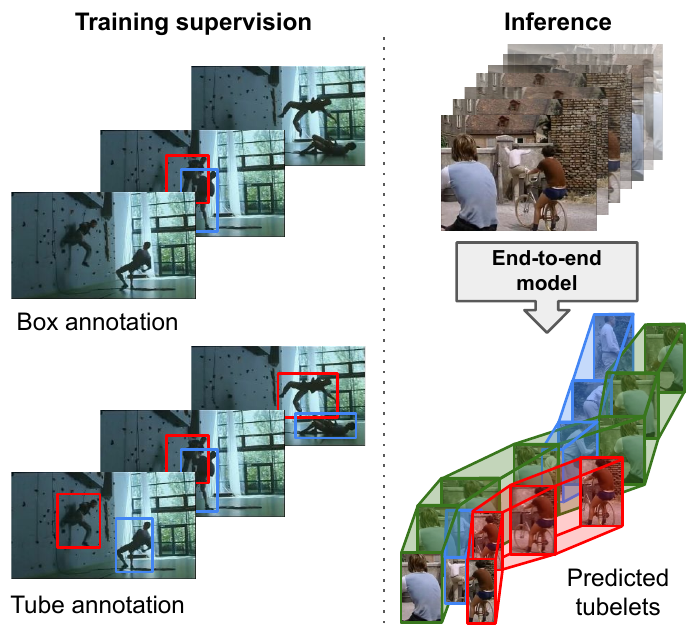}
	\label{fig:teaser}
	\vspace{-\baselineskip}
	\caption{
		We propose an end-to-end Spatio-Temporal Action Recognition model named STAR.
		Our model is end-to-end in that it does not require any external region proposals to predict \textit{tubelets} -- sequences of bounding boxes associated with a given person in every frame and their corresponding action classes.
	Our model can be trained with either sparse box annotations on selected keyframes, or full tubelet supervision. 
	}
	\vspace{-1.5\baselineskip}
\end{figure}

And whilst proposal-free algorithms, which do not require external person detectors, have been developed for detecting both at the keyframe-level~\cite{kopuklu2019you, chen2021watch, sun_eccv_2018} and tubelet-level~\cite{kalogeiton_iccv_2017, zhao2022tuber}, their performance has typically lagged behind their proposal-based counterparts. Here, we show for the first time that an end-to-end trainable spatio-temporal model outperforms a two-stage approach.

As shown in Fig.~\ref{fig:teaser}, we propose our {\bf S}patio-{\bf T}emporal {\bf A}ction Transforme{\bf R} (STAR) that consists of a pure-transformer architecture, and is based on the DETR~\cite{carion_eccv_2020} detection model.
Our model is ``end-to-end'' in that it does not require pre-processing in the form of proposals, nor post-processing in the form of non-maximal suppression (NMS) in contrast to the majority of prior work.
The initial stage of the model is a vision encoder.
This is followed by a decoder that processes learned latent queries, which represent each actor in the video, into output tubelets -- a sequence of bounding boxes and action classes at each time step of the input video clip.
Our model is versatile in that we can train it with either fully-labeled tube annotations, or with sparse keyframe annotations (when only a limited number of keyframes are labelled).
In the latter case, our network still predicts tubelets, and learns to associate detections of an actor, from one frame to the next, without explicit supervision.
This behaviour is facilitated by our formulation of factorised queries, decoder architecture and tubelet matching in the loss which all contain temporal inductive biases.

We conduct thorough ablation studies of these modelling choices. %
Informed by these experiments, we achieve state-of-the-art on both keyframe-based action localisation datasets like AVA~\cite{gu_cvpr_2018} and AVA-Kinetics~\cite{li2020ava}, and also tubelet-based datasets like UCF101-24~\cite{soomro_arxiv_2012} and JHMDB~\cite{jhuang2013towards}.
In particular, we achieve a Frame mAP of 44.6 on AVA-Kinetics outperforming previous published work~\cite{pan2021actor} by 8.2 points, and a recent foundation model~\cite{wang2022internvideo} by 2.1 points. In addition, our Video AP50 on UCF101-24 surpasses prior work~\cite{zhao2022tuber} by 11.6 points.
Moreover, our state-of-the-art results are achieved with a single forward-pass through the model, using only a video clip as input, and without any separate external person detectors providing proposals~\cite{wu2022memvit,wang2022internvideo, arnab2022beyond}, complex memory banks~\cite{wu2022memvit,zhao2022tuber,pan2021actor}, or additional object detectors~\cite{tang2020asynchronous,arnab2021unified}, as used by the prior state-of-the-art.
Furthermore, we outperform these complex, prior, state-of-the-art two-stage models whilst also having additional functionality in that our model predicts tubelets, that is, temporally consistent bounding boxes at each frame of the input video clip. %

\section{Related Work}

Models for spatio-temporal action localisation have typically built upon advances in object detectors for images.
The most performant methods for action localisation~\cite{pan2021actor,wu2022memvit,tang2020asynchronous,arnab2022beyond,feichtenhofer_iccv_2019} are based on ``two-stage'' detectors like Fast-RCNN~\cite{girshick_iccv_2015}. %
These models use external, pre-computed person detections, and use them to ROI-pool features which are then classified into action classes.
Although these models are cumbersome in that they require an additional model and backbone to first detect people, and therefore additional detection training data as well, they are currently the leading approaches on datasets such as AVA~\cite{gu_cvpr_2018}.
Such models using external proposals are also particularly suited to datasets such as AVA~\cite{gu_cvpr_2018} as each person is exhaustively labelled as performing an action, and therefore there are fewer false-positives from using action-agnostic person detections compared to datasets such as UCF101~\cite{soomro_arxiv_2012}.

The performance of these two-stage models has further been improved by incorporating more contextual information using feature banks extracted from additional frames in the video~\cite{wu2022memvit, pan2021actor, tang2020asynchronous, wu_cvpr_2019} or by utilising detections of additional objects in the scene~\cite{arnab2021unified, baradel_eccv_2018, zhang_tokmakov_cvpr_2019, wang_eccv_2018}.
Both of these cases require significant additional computation and complexity to train additional auxiliary models and to precompute features from them that are then used during training and inference of the localisation model.

Our proposed method, in contrast, is end-to-end in that it directly produces detections without any additional inputs besides a video clip.
Moreover, it outperforms these prior works without resorting to external proposals or memory banks, showing that a transformer backbone is sufficient to capture long-range dependencies in the input video.
In addition, unlike previous two-stage methods, our method directly predicts tubelets: a sequence of bounding boxes and actions for each frame of the input video, and can do so even when we do not have full tubelet annotations available.

A number of proposal-free action localisation models have also been developed~\cite{kopuklu2019you, chen2021watch, sun_eccv_2018, girdhar_cvpr_2019, kalogeiton_iccv_2017, zhao2022tuber}.
These methods are based upon alternative object detection architectures such as SSD~\cite{liu_ssd_eccv_2016}, CentreNet~\cite{zhou2019objects}, YOLO~\cite{redmon2016you}, DETR~\cite{carion_eccv_2020} and Sparse-RCNN~\cite{sun2021sparse}.
However, in contrast to our approach, they have been outperformed by their proposal-based counterparts.
Moreover, some of these methods~\cite{kopuklu2019you, girdhar_cvpr_2019, sun_eccv_2018} also consist of separate network backbones for learning video feature representations and proposals for a keyframe, and are thus effectively two networks trained jointly, and cannot predict tubelets either. %

Among prior works that do not use external proposals, and also directly predict tubelets~\cite{kalogeiton_iccv_2017,li2020actions,song2019tacnet,li2018recurrent,singh_iccv_2017,zhao2022tuber}, our work is the most similar to TubeR~\cite{zhao2022tuber} given that our model is also based on DETR.
Our model, however, is purely transformer-based (including the encoder) and achieves substantially higher performance without requiring external memory banks precomputed offline like~\cite{zhao2022tuber}.
Furthermore, unlike TubeR, we also demonstrate how our model can predict tubelets (\ie predictions at every frame of the input video), even when we only have sparse keyframe supervision (\ie ground truth annotation for a limited number of frames) available. %

Finally, we note that DETR has also been extended as a proposal-free method to addressing different localisation tasks in video such as video instance segmentation~\cite{wang2021end}, temporal localisation~\cite{liu2022end,zhang2021temporal,nawhal2021activity} and moment retrieval~\cite{lei2021detecting}. %

\section{Spatio-Temporal Action Transformer}
\label{sec:method}

\begin{figure*}[t]
	\vspace{-1.5\baselineskip}
	\includegraphics[width=\linewidth, trim={0 8.6cm 1.2cm 0}, clip]{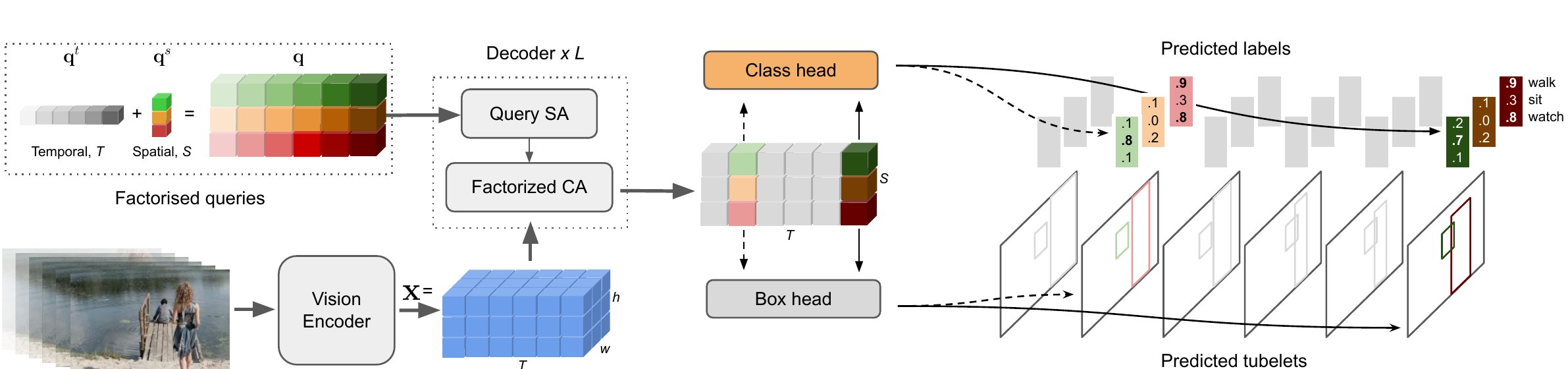}
	\caption{
	Our model processes a fixed-length video clip, and for each frame, outputs tubelets (\ie linked bounding boxes with associated action class probabilities).
	It consists of a transformer-based vision encoder which outputs a video representation, $\mathbf{x} \in \mathbb{R}^{T \times h \times w \times d}$.
	The video representation, along with learned queries, $\mathbf{q}$ (which are factorised into spatial $\mathbf{q}^s$ and temporal components $\mathbf{q}^t$) are decoded into tubelets by a decoder of $L$ layers followed by shallow box and class prediction heads.
	}
 	\vspace{-\baselineskip}
	\label{fig:model_overview}
\end{figure*}

Our proposed model ingests a sequence of video frames, and directly predicts tubelets (a sequence of bounding boxes and action labels). %
In contrast to leading spatio-temporal action recognition models, our model does not use external person detections~\cite{pan2021actor, arnab2022beyond, tong2022videomae, wu2022memvit} or external memory banks~\cite{wu_cvpr_2019, zhao2022tuber, pan2021actor} to achieve strong results.%

As summarised in Fig.~\ref{fig:model_overview}, our model consists of a vision encoder (Sec.~\ref{sec:method_vision_encoder}), followed by a decoder which processes learned query tokens into output tubelets (Sec.~\ref{sec:method_decoder}).
We incorporate temporal inductive biases into our decoder to improve accuracy and tubelet prediction with weaker supervision. %
Our model is inspired by the DETR architecture~\cite{carion_eccv_2020} for object detection in images, and is also trained with a set-based loss and Hungarian matching.
We detail our loss, and how we can train with either sparse keyframe supervision or full tubelet supervision, in Sec.~\ref{sec:method_loss}.

\subsection{Vision Encoder}
\label{sec:method_vision_encoder}

The vision backbone processes an input video, $\mathbf{X} \in \mathbb{R}^{T \times H \times W \times 3}$ to produce a feature representation of the input video $\mathbf{x} \in \mathbb{R}^{t \times h \times w \times d}$.
Here, $T$, $H$ and $W$ are the original temporal-, height- and width-dimensions of the input video respectively, whilst $t$, $h$ and $w$ are the spatio-temporal dimensions of their feature representation, and $d$ its latent dimension.
As we use a transformer, specifically the ViViT Factorised Encoder~\cite{arnab2021vivit}, these spatio-temporal dimensions depend on the patch 
size when tokenising the input.
To retain spatio-temporal information, we remove the spatial- and temporal-aggregation steps in the original transformer backbone.
And if the temporal patch size is larger than 1, we bilinearly upsample the final feature map along the temporal axis to maintain the original temporal resolution.

\subsection{Tubelet Decoder}
\label{sec:method_decoder}

Our decoder processes the visual features, $\mathbf{x} \in \mathbb{R}^{T \times h \times w \times c}$, along with learned queries, $\mathbf{q} \in \mathbb{R}^{T \times S \times d}$, to outputs tubelets, $\mathbf{y} = (\mathbf{b}, \mathbf{a})$ which are a sequence of bounding boxes, $b \in \mathbb{R}^{T \times S \times 4}$ and corresponding actions, $a \in \mathbb{R}^{T \times S \times C}$.
Here, $S$ denotes the maximum number of bounding boxes per frame (padded with ``background'' as necessary) and $C$ denotes the number of output classes. %

The idea of decoding learned queries into output detections using the transformer decoder architecture of Vaswani~\etal~\cite{vaswani_neurips_2017} was used in DETR~\cite{carion_eccv_2020}. %
In summary, the decoder of~\cite{carion_eccv_2020, vaswani_neurips_2017} consists of $L$ layers, each performing a series of self-attention operations on the queries, and cross-attention between the queries and encoder outputs.

\begin{figure}[t]
	\vspace{-0.5\baselineskip}
	\includegraphics[width=\linewidth, trim={0 10.1cm 18.2cm 0}, clip]{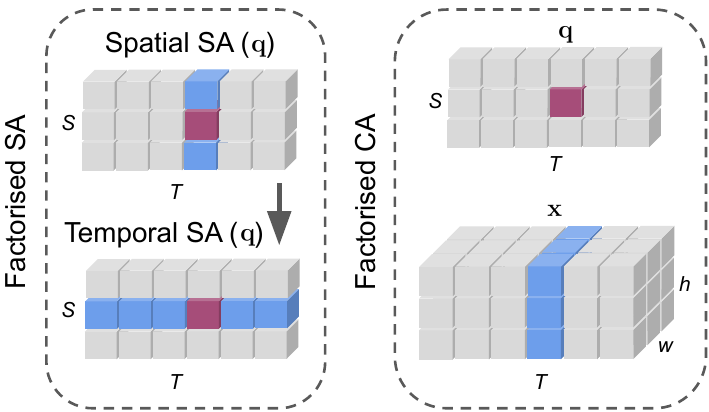}
	\caption{
		Our decoder layer consists of factorised self-attention (SA) (left) and cross-attention (CA) (right) operations designed to provide a spatio-temporal inductive bias and reduce computation. %
		Both operations restrict attention to the same spatial and temporal slices as the query token, as illustrated by the receptive field (blue) for a given query token (magenta).
		Factorised SA consists of two operations, whilst in Factorised CA, there is one operation.
	}
	\vspace{-\baselineskip}
	\label{fig:model_decoder}
\end{figure}

We modify the queries, self-attention and cross-attention operations for our spatio-temporal localisation scenario, as shown in Fig.~\ref{fig:model_overview} and~\ref{fig:model_decoder} to include additional temporal inductive biases, and to improve accuracy as detailed below. 

\vspace{\paravspace}
\paragraph{Queries}
\label{sec:method_queries}

Queries, $\mathbf{q}$, in DETR, are decoded using the encoded visual features, $\mathbf{x}$, into bounding box predictions, and are analogous to the ``anchors'' used in other detection architectures such as Faster-RCNN~\cite{ren_neurips_2015}.

The most straightforward way to define queries is to randomly initialise $\mathbf{q} \in \mathbb{R}^{T \times S \times d}$, where there are $S$ bounding boxes at each of the $T$ input frames in the video clip.

However, we find it is more effective to factorise the queries into separate learned spatial, $\mathbf{q}^s \in \mathbb{R}^{S \times d}$, and temporal, $\mathbf{q}^{T \times d}$ parameters.
To obtain the final tubelet queries, we simply repeat the spatial queries across all frames, and add them to their corresponding temporal embedding at each location, as shown in Fig.~\ref{fig:model_overview}.
More concretely $\mathbf{q}_{ij} = \mathbf{q}^t_i + \mathbf{q}^s_j$ where $i$ and $j$ denote the temporal and spatial indices respectively.

The factorised query representation means that the same spatial embedding is used across all frames.
Intuitively, this encourages the $i^{th}$ spatial query embedding, $\mathbf{q}^s_i$, to bind to the same location across different frames of the video, and since objects typically have small displacements from frame to frame, may help to associate bounding boxes within a tubelet together. %
We verify this intuition empirically in the experimental section. 

\vspace{\paravspace}
\paragraph{Decoder layer}

The decoder layer in the original transformer~\cite{vaswani_neurips_2017} consists of self-attention on the queries, $\mathbf{q}$, followed by cross-attention between the queries and the outputs of the encoder, $\mathbf{x}$, and then a multilayer perceptron (MLP) layer~\cite{vaswani_neurips_2017, hendrycks_arxiv_2016}.
These operations can be denoted by
\begin{align}
	\mathbf{u}^{\ell} &= \text{MHSA}(\mathbf{q}^{\ell}) + \mathbf{q}^{\ell}, \\
	\mathbf{v}^{\ell} &= \text{CA}(\mathbf{u}^{\ell}, \mathbf{x}) + \mathbf{u}^{\ell}, \\
	\mathbf{z}^{\ell} &= \text{MLP}(\mathbf{v}^{\ell}) + \mathbf{v}^{\ell},
\end{align}
where $\mathbf{z}^{\ell}$ is the output of the $\ell^{th}$ decoder layer, $\mathbf{u}$ and $\mathbf{v}$ are intermediate variables, %
$\text{MHSA}$ denotes multi-headed self-attention and $\text{CA}$ cross-attention.  %
Note that the inputs to the MLP and self- and cross-attention operations are layer-normalised~\cite{ba_arxiv_2016}, which we omit here for clarity. %

In our model, we factorise the self- and cross-attention layers across space and time respectively as shown in Fig.~\ref{fig:model_decoder}, to introduce a temporal locality inductive bias, and also to increase model efficiency.
Concretely, when applying $\text{MHSA}$, we first compute the queries, keys and values, over which we attend twice: first independently at each time step with each frame, and then, independently along the time axis at each spatial location.
Similarly, we modify the cross-attention operation so that only tubelet queries and backbone features from the same time index attend to each other.

\vspace{\paravspace}
\paragraph{Localisation and classification heads}

We obtain the final predictions of the network, $\mathbf{y} = (\mathbf{b}, \mathbf{a})$, by applying a small feed-forward network to the outputs to the decoder, $\mathbf{z}$, following DETR~\cite{carion_eccv_2020}. %
The sequence of bounding boxes, $\mathbf{b}$, is obtained with a 3-layer MLP, and is parameterised by the box center, width and height for each frame in the tubelet.
A single-layer linear projection is used to obtain class logits, $\mathbf{a}$.
As we predict a fixed number of $S$  bounding boxes per frame, and $S$ is more than the maximum number
of ground truth instances in the frame, we also include an additional class label, $\varnothing$, which represents the ``background'' class 
 which tubelets with no action class can be assigned to.

\subsection{Training objective}
\label{sec:method_loss}

Our model predicts bounding boxes and action classes at each frame of the input video.
Many datasets, however, such as AVA~\cite{gu_cvpr_2018}, are only sparsely annotated at selected keyframes of the video.
In order to leverage the available annotations, we compute our training loss, Eq.~\ref{eq:training_loss}, only at the annotated frames of the video, after having matched the predictions to the ground truth.
This is denoted as %
\begin{equation}
	\mathcal{L}(\mathbf{y}, \hat{\mathbf{y}}) = \frac{1}{|\mathcal{T}|} \sum_{t \in \mathcal{T}} \mathcal{L}_{\text{frame}}(\mathbf{y}, \hat{\mathbf{y}}) \label{eq:training_loss},
\end{equation}
where $\mathcal{T}$ is the set of labelled frames; $\mathbf{y}$ and $\mathbf{\hat{y}}$ denote the ground truth and predicted tubelets after matching. %

Following DETR~\cite{carion_eccv_2020}, our training loss at each frame, $\mathcal{L}_{\text{frame}}$, is a sum of an $L_1$ regression loss on bounding boxes, the generalised IoU loss~\cite{rezatofighi2019generalized} on bounding boxes, and a cross-entropy loss on action labels:
\begin{align}
	\mathcal{L}_{\text{frame}}(\mathbf{b}^t, \hat{\mathbf{b}}^t, \mathbf{a}^t, \hat{\mathbf{a}}^t) &= \sum_{i} \mathcal{L}_{\text{box}}(\mathbf{b}^t_i, \hat{\mathbf{b}}^t_i) +  \mathcal{L}_{\text{iou}}(\mathbf{b}^t_i, \hat{\mathbf{b}}^t_i) \nonumber \\
	& \quad \qquad + \mathcal{L}_{\text{class}}(\mathbf{a}^t_i, \hat{\mathbf{a}}^t_i).
	\label{eq:frame_loss}
\end{align}
\paragraph{Matching}
Set-based detection models such as DETR can make predictions in any order, which is why the predictions need to be matched to the ground truth before computing the training loss. %

The first form of matching that we consider is to independently perform bipartite matching at each frame to align the model's predictions to the ground truth (or the $\varnothing$ background class) before computing the loss.
In this case, we use the Hungarian algorithm~\cite{kuhn1955hungarian} to obtain $T$ %
permutations of $S$ elements, $\hat{\pi}^t \in \Pi^t$, at each frame, where the permutation at the $t^{th}$ frame minimises the per-frame loss,
\begin{align}
	\hat{\pi}^t = \argmin_{\pi \in \Pi^{t}} \mathcal{L}_{\text{frame}}(\mathbf{y}^t, \hat{\mathbf{y}}^t_{\pi(i)}).
\end{align}
An alternative is to perform \textit{tubelet matching}, where all queries with the same spatial index, $\mathbf{q}^s$, must match to the same ground truth annotation across all frames of the input video.
Here the permutation is obtained over $S$ elements as
\begin{align}
	\hat{\pi} = \argmin_{\pi \in \Pi} \frac{1}{|\mathcal{T}|}\sum_{t \in \mathcal{T}} \mathcal{L}_{\text{frame}}(\mathbf{y}^t, \hat{\mathbf{y}}^t_{\pi^{t}(i)}).
\end{align}
Intuitively, tubelet matching provides stronger supervision when we have full tubelet annotations available. %
Note that regardless of the type of matching that we perform, the loss computation and the overall model architecture remains the same.
Note that we do not weight terms in Eq.~\ref{eq:frame_loss}, for both matching and loss calculation, for simplicity, and to avoid having additional hyperparameters, as also done in~\cite{minderer2022simple}.

\subsection{Discussion}
\label{sec:method_discussion}

As our approach is based on DETR, it does not require external proposals nor non-maximal suppression for post-processing.
The idea of using DETR for action localisation has also been explored by TubeR~\cite{zhao2022tuber} and WOO~\cite{chen2021watch}.
There are, however, a number of key differences:
WOO does not detect tubelets at all, but only actions at the center keyframe. %
We also factorise our queries in the spatial and temporal dimensions (Sec.~\ref{sec:method_queries}) to provide inductive biases urging spatio-temporal association. %
Moreover, we predict action classes separately for each time step in the tubelet, meaning that each of our queries binds to an actor in the video. %
TubeR, in contrast, parameterises queries such that they are each associated with separate actions (features are average-pooled over the tubelet, and then linearly classified into a single action class).
This choice also means that TubeR requires an additional ``action switch'' head to predict when tubelets start and end, which we do not require as different time steps in a tubelet can have different action classes in our model.
Furthermore, we show experimentally (Tab.~\ref{tab:ablation_detection_architectures}) that TubeR's parameterisation obtains lower accuracy.
We also consider two types of matching in the loss computation (Sec.~\ref{sec:method_loss}) unlike TubeR, with ``tubelet matching'' designed for predicting more temporally consistent tubelets.
And in contrast to TubeR, we experimentally show how our decoder design allows our model to accurately predict tubelets even with weak, keyframe supervision. %

Finally, TubeR requires additional complexity in the form of a ``short-term context module''~\cite{zhao2022tuber} and the external memory bank of~\cite{wu_cvpr_2019} which is computed offline using a separate model to achieve strong results.
As we show experimentally in the next section, we outperform TubeR without any additional modules, meaning that our model does indeed produce tubelets in an end-to-end manner.

\section{Experimental Evaluation}

\subsection{Experimental set-up}

\paragraph{Datasets}
We evaluate on four spatio-temporal action localisation benchmarks.
AVA and AVA-Kinetics contain sparse annotations at each keyframe, %
whereas UCF101-24 and JHMDB51-21 contain full tubelet annotations.

\textit{AVA~\cite{gu_cvpr_2018}} consists of 430, 15-minute video clips from movies.
Keyframes are annotated at every second in the video, with about 210 000 labelled frames in the training set, and 57 000 in the validation set.
There are 80 atomic actions labelled for every actor in the clip, of which 60 are used for evaluation~\cite{gu_cvpr_2018}. 
Following standard practice, we report the Frame~Average Precision~(fAP) at an IoU threshold of 0.5 using the latest v2.2 annotations~\cite{gu_cvpr_2018}.

\textit{AVA-Kinetics~\cite{li2020ava}} is a superset of AVA, and adds detection annotations following the AVA protocol, to a subset of Kinetics 700~\cite{carreira2019short} videos.
Only a single keyframe in a 10-second Kinetics clip is labelled.
In total, about 140 000 labelled keyframes are added to the training set, and 32 000 to the validation sets of AVA. 
Once again, we follow standard practice in reporting the Frame AP at a 0.5 IoU threshold. %

\textit{UCF101-24~\cite{soomro_arxiv_2012}} is a subset of UCF101, and annotates 24 action classes 
with full spatio-temporal tubes in 3 207 untrimmed videos.
Note that actions are not labelled exhaustively as in AVA, and there may be people present in the video who are not performing any labelled action.
Following standard practice, we use the corrected annotations of ~\cite{singh_iccv_2017}.
We report both the Frame AP, which evaluates the predictions at each frame independently, and also the Video AP.
The Video AP uses a 3D, spatio-temporal IoU to match predictions to targets.
And since UCF101-24 videos are up to 900 frames long (median length of 164 frames), and our network processes $T=32$ frames at a time, we link together tubelet predictions from our network into full-video-tubes using the same causal linking algorithm as~\cite{kalogeiton_iccv_2017, li2020actions} for fair comparison.

\textit{JHMDB51-21~\cite{jhuang2013towards}} also contains full tube annotations in 928 trimmed videos.
However, as the videos are shorter and at most 40 frames, we can process the entire clip with our network, and do not need to perform any linking.

\vspace{\paravspace}
\paragraph{Implementation details}

For our vision encoder backbone, we use ViViT Factorised Encoder~\cite{arnab2021vivit} where model sizes, such as ``Base'' and ``Large'' follow the original definitions from~\cite{devlin_naacl_2019,dosovitskiy_iclr_2021}.
It is initialised from pretrained checkpoints, which are typically first pretrained on image datasets like ImageNet-21K~\cite{deng_cvpr_2009} and then finetuned on video datasets like Kinetics~\cite{kay_arxiv_2017}.
Our model processes $T = 32$ unless otherwise specified, and has $S = 64$ spatial queries per frame, and the latent dimensionality of the decoder is $d=2048$.
Exhaustive implementation details and training hyperparameters are included in the supplementary.

\subsection{Ablation studies}

We analyse the design choices in our model by conducting experiments on both AVA (with sparse per-frame supervision) and on UCF101-24 (where we can evaluate the quality of our predicted tubelets).
Unless otherwise stated, our backbone is ViViT-Base pretrained on Kinetics 400, and the frame resolution is 160 pixels (160p) on the smaller side. %

\vspace{\paravspace}
\paragraph{Comparison of detection architectures}
\begin{table}[t]
\centering
\caption{
    Comparison of detection architectures on AVA controlling for the same backbone (ViViT-B), resolution (160p) and training settings.
    Our end-to-end approach outperforms proposal-based ROI models.
    Binding each query to a person, rather than to an action (as done in TubeR~\cite{zhao2022tuber}), also yields solid improvements.
}
\begin{tabular}{lcc}
    \toprule
                          & Proposals    & AP50 \\ 
    \midrule
    Two-stage ROI model   &  \cite{wu_cvpr_2019}            &   $25.2$   \\ %
    Query binds to action &  None            &   $23.6$  \\  %
    Ours, query binds to person &  None            &   $\mathbf{26.7}$   \\ %
    \bottomrule
\end{tabular}
\label{tab:ablation_detection_architectures}
\vspace{-\baselineskip}
\end{table}

Table~\ref{tab:ablation_detection_architectures} compares our model to two relevant baselines: Firstly, a two-stage Fast-RCNN model using external person detections from~\cite{wu_cvpr_2019} (as used by~\cite{feichtenhofer_iccv_2019,fan2021multiscale,arnab2022beyond,wu_cvpr_2019}).
And secondly, we compare to using the query parameterisation used in TubeR~\cite{zhao2022tuber}, where each query binds to an action, as described in Sec.~\ref{sec:method_discussion}.
We control other experimental settings by using the same backbone (ViViT-Base) and resolution (160p).

The first row of Tab.~\ref{tab:ablation_detection_architectures} shows that our end-to-end model improves upon a two-stage model by 1.5 points on AVA, emphasising the promise of our approach.
Note that the proposals of~\cite{wu_cvpr_2019} achieve an AP50 of 93.9 for person detection on the AVA validation set.
They were obtained by first pretraining a Faster-RCNN~\cite{ren_neurips_2015} detector on COCO keypoints, and then finetuning on the person boxes from the training set of AVA, using a resolution of 1333 on the longer side.
Our model is end-to-end, and does not require external proposals generated by a separate model at all.

The second row of Tab.~\ref{tab:ablation_detection_architectures} compares our model, where each query represents a person and all of their actions (Sec.~\ref{sec:method_decoder}) to the approach of TubeR~\cite{zhao2022tuber} (Sec.~\ref{sec:method_discussion}), where there is a separate query for each action being performed.
We observe that this parameterisation has a substantial impact, with our method outperforming it significantly by 3.1 points, motivating the design of our decoder.

\vspace{\paravspace}
\paragraph{Query parameterisation}

\begin{table}
    \centering
    \caption{Comparison of independent and factorised queries on the AVA and UCF101-24 datasets. Factorised queries are particularly beneficial for predicting tubelets, as shown by the VideoAP on UCF101-24 which has full tube annotations. Both models use tubelet matching in the loss.
	}
    \scalebox{0.89}{
        \begin{tabular}{lccccc}
        \toprule
            & AVA      & \multicolumn{4}{c}{UCF101-24} \\
        
        \cmidrule(lr){2-2} \cmidrule(lr){3-6}
        
        Query  &  fAP & fAP  &  vAP20 &     vAP50 &   vAP50:95  \\
        
        \midrule      
        Independent  &  $25.2$  &  $85.6$           &  $86.3$  &  $59.5$  &  $28.9$  \\  %
        Factorised   &  $\mathbf{26.3}$  &  $\mathbf{86.5}$  &  $\mathbf{87.4}$  &  $\mathbf{63.4}$  &  $\mathbf{29.8}$   \\  %
        \bottomrule
        \end{tabular}
    }
	\vspace{-0.5\baselineskip}
\label{tab:ablation_queries}
\end{table}

\begin{table}
    \centering
    \caption{Comparison of independent and tubelet matching for computing the loss on AVA and UCF101-24.
    Tubelet matching helps for tube-level evaluation metrics like the Video AP (vAP) on UCF101-24.
    Note that tubelet matching is actually still possible on AVA as the annotations are at 1fps with actor identities.}
    \scalebox{0.815}{
        \begin{tabular}{lccccc}
        \toprule
            & AVA      & \multicolumn{4}{c}{UCF101-24} \\
        
        \cmidrule(lr){2-2} \cmidrule(lr){3-6}
        
        Query  &  fAP & fAP &  vAP20 &  vAP50 &   vAP50:95 \\
        
        \midrule      
        Per-frame matching  &  $\mathbf{26.7}$ & $\mathbf{88.2}$   &  $85.7$  &  $63.5$  &  $29.4$ \\
        Tubelet matching   &  $26.3$  &  $86.5$  &  $\mathbf{87.4}$  &  $63.4$  &  $\mathbf{29.8}$  \\  %
        \bottomrule
        \end{tabular}
        }
\label{tab:ablation_matching_loss}
\vspace{-\baselineskip}
\end{table}

Table~\ref{tab:ablation_queries} compares our independent and factorised query methods (Sec.~\ref{sec:method_queries}) on AVA and UCF101-24.
We observe that factorised queries consistently provide improvements on both the Frame AP and the Video AP across both datasets.
As hypothesised in Sec.~\ref{sec:method_queries}, we believe that this is due to the inductive bias present in this parameterisation.
Note that we can only measure the Video AP on UCF101-24 as it has tubes labelled.

\vspace{\paravspace}
\paragraph{Matching for loss calculation}
As described in Sec.~\ref{sec:method_loss}, when matching the predictions to the ground truth for loss computation, we can either independently match the outputs at each frame to the ground truths at each frame, or, we can match the entire predicted tubelets to the ground truth tubelets.
Table~\ref{tab:ablation_matching_loss} shows that tubelet matching does indeed improve the quality of the predicted tubelets, as shown by the Video AP on UCF101-24.
However, this comes at the cost of the quality of per-frame predictions (\ie Frame AP).
This suggests that tubelet matching improves the association of bounding boxes predicted at different frames (hence higher Video AP), but may also impair the quality of the bounding boxes predicted at each frame (Frame AP).
Note that it is technically possible for us to also perform tubelet matching on AVA, since AVA is annotated at 1fps with actor identities, and our model is input 32 frames at 12.5fps (therefore 2.56 seconds of temporal context) meaning that we have sparse tubelets with 2 or 3 annotated frames.

As tubelet matching helps with the overall Video AP, we use it for subsequent experiments on UCF101-24.
For AVA, we use per-frame matching as the standard evaluation metric is the Frame AP, and annotations are sparse at 1fps.

\vspace{\paravspace}
\paragraph{Weakly-supervised tubelet detection}
\begin{table}[t]
\caption{
Our model can predict tubelets even when the ground truth annotations are sparse.
We show this by subsampling training annotations from the UCF101-24 dataset.
Our model sees minimal performance deterioration even when using only $1 / 24$ or 4\% of the annotated frames.
}

\scalebox{0.76}{
    \begin{tabular}{lccccc}
    \toprule
    Sampling        & Labelled frames &  fAP   & vAP20  & vAP50  & vAP50:95 \\
    \midrule
    All frames & $458\,814$ &  $86.5$  & $87.4$  &  $63.4$  &  $29.8$  \\  %
    Every 12        &  \phantom{$0$}$39\,237$   &  $85.2$  &  $87.2$  &  $63.0$  &  $29.3$  \\  %
    Every 24        &  \phantom{$0$}$20\,243$   &  $84.9$  &  $86.8$  &  $63.2$  &  $28.1$  \\  %
    One per video   &  \phantom{$00$}$2\,284$    &  $70.2$  &  $77.1$  &  $48.5$  &  $20.4$  \\  %
    \bottomrule
    \end{tabular}
}
\label{tab:ablation_ucf_weak_supervision}
\vspace{-0.5\baselineskip}
\end{table}

Our model can predict tubelets even when the ground truth annotations are sparse and only labelled at certain frames (such as the AVA dataset).
We quantitatively measure this ability of our model on the UCF101-24 dataset which has full tube annotations.
We do so by subsampling labels from the training set, and evaluating the full tubes on the validation set.

As shown in Tab.~\ref{tab:ablation_ucf_weak_supervision}, we still obtain meaningful tube predictions, with a Video AP20 of 77.1, when using only a single frame of annotation from each UCF video clip.
When retaining 1 frame of supervision for every 24 labelled frames (which is roughly 1fps and corresponds to the AVA dataset's annotations), we observe minimal deterioration with respect to the fully supervised model (all Video AP metrics are within 0.7 points).
And retaining 1 frame of annotation for every 12 consecutive labelled frames performs similarly to using all frames in the video clip.
These results suggest that due to the redundancy in the dataset (motion between frames is often limited), and the inductive bias of our model, we do not require every frame in the tube to be labelled in order to produce accurate tubelet predictions.

\vspace{\paravspace}
\paragraph{Decoder design}

\begin{table}[t]
\centering
\caption{
	Effect of decoder depth on performance on the AVA dataset. %
	Performance saturates at $L = 6$ layers.
}
\begin{tabular}{lccccc}
\toprule
Layers ($L$) & 0 & 1 & 3 & 6 & 9 \\
\midrule
mAP~$\uparrow$ & $23.4$ & $24.6$ & $26.2$ & $26.5$ &  $\mathbf{26.7}$ \\	
\bottomrule
\end{tabular}
\vspace{-1\baselineskip}
\label{tab:ablation_decoder_layers}

\end{table}

\begin{table}[t]
\vspace{-0.2\baselineskip}
\centering
\caption{Effect of the type of attention used in the decoder on AVA.
	Factorised attention is both more accurate and efficient (almost half of the GFLOPs per decoder layer).}
\begin{tabular}{lcc}
\toprule
Decoder attention & mAP & GFLOPs \\
\midrule
Full & $26.4$ & $10.5$ \\
Factorised & $\mathbf{26.7}$ & $\mathbf{5.3}$  \\
\bottomrule
\end{tabular}
\label{tab:ablation_decoder_attention}
\vspace{-0.5\baselineskip}
\end{table}

Tables~\ref{tab:ablation_decoder_layers} and~\ref{tab:ablation_decoder_attention} analyse the effect of the decoder depth and the type of attention in the decoder (described in Sec.~\ref{sec:method_decoder}).
As seen in Tab.~\ref{tab:ablation_decoder_layers}, detection accuracy on AVA increases with the number of decoder layers, plateauing at around 6 layers. %
It is possible to use no decoder layers too: In this case, instead of learning queries $\mathbf{q}$ (Sec.~\ref{sec:method_queries}), we simply interpret the outputs of the vision encoder (Sec.~\ref{sec:method_vision_encoder}), $\mathbf{x}$, as our queries %
and apply the localisation and classification heads directly upon them.
Using decoder layers, however, can provide a performance increase of up to 3.3 mAP points (14\% relative), emphasising their utility. 

Table~\ref{tab:ablation_decoder_attention} shows that factorised attention in the decoder is more accurate than standard, ``full'' attention between all queries and visual features.
Moreover, it is more efficient too, using almost half of the GFLOPs at each decoder layer.

\vspace{\paravspace}
\paragraph{Effect of resolution and pretraining}

\begin{table}[t]
    \centering
    \caption{
    Increasing the image resolution on the AVA dataset leads to consistent accuracy improvements, primarily on small objects.
    APs, APm and APl denote the AP at 0.5 IoU threshold on small, medium and large boxes respectively following the COCO protocol~\cite{lin_coco_eccv_2014}.
    AVA videos have a median aspect ratio of 16:10, and we pad the larger side when the aspect ratio is different.
    }
    \begin{tabular}{lcccc}
    \toprule
    Resolution                 & mAP  & APs & APm & APl \\  %
    \midrule
    140 $\times$ 224           &  $25.4$ & $7.2$ & $11.2$  & $27.8$ \\
    160 $\times$ 256           &  $26.7$ & $11.5$ & $12.5$ & $28.7$ \\ %
    220 $\times$ 352           &  $28.8$ & $12.0$ & $15.1$ & $30.7$ \\
    260 $\times$ 416           &  $29.4$ & $13.3$ & $15.8$ & $31.0$ \\
    320 $\times$ 512           &  $30.0$ & $17.5$ & $16.0$ & $32.0$ \\  %
    \bottomrule
    \end{tabular}
    \label{tab:ablation_resolution}
    \vspace{-0.5\baselineskip}
\end{table}
    
\begin{table}[t]
\centering
\caption{
	Comparison of pretraining for our models with ViViT-B and ViViT-L backbones on AVA using a resolution of $160 \times 256$.
    Larger models benefit more from additional initial pretraining.
}
\begin{tabular}{lcc}
\toprule
Pretrain                 & STAR/B & STAR/L \\

\midrule

IN21K~\cite{deng_cvpr_2009} $\to$ K400~\cite{kay_arxiv_2017}         &   $26.7$  &  $27.0$  \\  %
IN21K~\cite{deng_cvpr_2009} $\to$ K700~\cite{carreira2019short}         &  $27.3$  &  $27.6$   \\  %
JFT~\cite{sun_iccv_2017} $\to$ WTS~\cite{stroud2020learning}          &   $\mathbf{31.1}$  &  $34.2$  \\  %
CLIP~\cite{radford2021learning} $\to$ K700~\cite{carreira2019short}          &  $30.3$  &  $\mathbf{36.2}$  \\  %

\bottomrule
\end{tabular}
\label{tab:ablation_pretraining}
\vspace{-1.25\baselineskip}
\end{table}

Scaling up the image resolution is critical to achieving high performance for object detection in images~\cite{huang2017speed, singh2018analysis}.
However, we are not aware of previous works studying this for video action localisation.
Table~\ref{tab:ablation_resolution} shows that we do indeed observe substantial improvements from higher resolution, improving by up to 4.6 points on AVA.
As expected, higher resolutions help more for detection at small sizes, where we follow the COCO~\cite{lin_coco_eccv_2014} convention of object sizes.
Note that AVA videos have a median aspect ratio of 16:10, and we pad the larger side for videos with different aspect ratios.

Similarly, Tab.~\ref{tab:ablation_pretraining} shows the effect of different pretraining datasets.
Video vision transformers are typically pretrained on an image dataset (like ImageNet-21K~\cite{deng_cvpr_2009} or JFT~\cite{sun_iccv_2017}), before being finetuned on a video dataset, such as Kinetics~\cite{kay_arxiv_2017}.
We find that the initial image checkpoint plays an important rule, with CLIP~\cite{radford2021learning} pretraining significantly outperforming supervised pretraining on ImageNet-21K~\cite{dosovitskiy_iclr_2021,steiner2022train}.
And perhaps surprisingly, we find that CLIP-Large-pretrained models, then finetuned on Kinetics 700~\cite{carreira2019short} outperform models pretrained on JFT~\cite{sun_iccv_2017} and finetuned on the large-scale, but noisy WTS dataset~\cite{stroud2020learning} of web-scraped videos.
Moreover, we find that large backbones benefit more from more pretraining data.

\subsection{Comparison to state-of-the-art}

We compare our model to the state-of-the-art on datasets with both sparsely annotated keyframes (AVA and AVA-Kinetics), and full tubes (UCF101-24 and JHMDB).

\vspace{\paravspace}
\paragraph{AVA and AVA-Kinetics}

\begin{table*}[t]
	\vspace{-0.5\baselineskip}
	\centering
	\caption{Comparison to the state-of-the-art (reported with mean Average Precision; mAP~$\uparrow$) on AVA~\cite{gu_cvpr_2018} and AVA-Kinetics~(AVA-K)~\cite{li2020ava}.
		For AVA, we use the latest v2.2 annotations.
		Methods using external proposals (i.e. not end-to-end) are also trained on additional object detection and human pose data.
		Unless otherwise stated, separate models are trained for AVA and AVA-Kinetics.
		$^{*}$ denotes the model was trained on AVA-Kinetics and evaluated on AVA.
		``Res.'' denotes the frame resolution of the shorter side.
	}
	\renewcommand{\arraystretch}{0.92}
	\scalebox{0.9}{
	\begin{tabular}{llcC{1.3cm}C{2cm}ccc}
		\toprule
		& Pretraining & Views & AVA & AVA-K & Res. & Backbone & End-to-end \\ 
		\midrule
		MViT-B~\cite{fan2021multiscale}   & K400 & 1 & $27.3$ & -- & -- & MViT & \xmark \\ 
		Unified~\cite{arnab2021unified} & K400 & 6 & $27.7$ & -- & 320 & SlowFast & \xmark \\

		AIA~\cite{tang2020asynchronous}  & K700     &    18        & $32.3$   & -- & 320 & SlowFast & \xmark \\
		ACAR~\cite{pan2021actor}  & K700     &       6     & $33.3$  & $36.4$  & 320 & SlowFast & \xmark  \\ 
		
		MeMViT~\cite{wu2022memvit} & K700 & -- & $34.4$ & -- & 312 & MViT v2 & \xmark \\ 
		Co-finetuning~\cite{arnab2022beyond} & IN21K$\to$K700, MiT, SSv2 & 1 & $32.8$ & $33.1$ & 320 & ViViT/L & \xmark \\
		& JFT,WTS$\to$K700, MiT, SSv2 & 1 & $36.1$ & $36.2$ & 320 & ViViT/L & \xmark \\  %
		VideoMAE~\cite{tong2022videomae} & SSL K700 $\to$ Sup. K700. & -- & $39.3$ & -- & 256 & ViViT/L & \xmark \\
		InternVideo$^{*}$~\cite{wang2022internvideo} & 7 different datasets & -- & $41.0$ & $42.5$ & -- & Uniformer v2 & \xmark \\
		\midrule
		Action Transformer~\cite{li2020ava} & K400 & 1 & -- & $23.0$ & 400 & I3D & \cmark \\
		WOO~\cite{chen2021watch} & K600 & 1 & $28.3$ & -- & 320 & SlowFast & \cmark \\
		TubeR~\cite{zhao2022tuber} & Instagram65M~\cite{mahajan_eccv_2018}$\to$K400 & 2 & $33.6$ & -- & 256 & CSN-152 & \cmark \vspace{0.3ex} \\
		STAR/B (ours) & IN21K$\to$K400 & 1 &  $30.0$ & $36.6$ & 320 & ViViT/B & \cmark \\  %
		& JFT$\to$WTS & 1 &  $36.3$ & $41.8$ & 320 & ViViT/B & \cmark \\  %
		& CLIP$\to$K700 & 1 &  $33.9$ & $39.1$ & 320 & ViViT/B & \cmark \\  %
		STAR/L (ours)    & JFT$\to$WTS & 1 &  $39.0$ & $\mathbf{44.6}$ & 320 & ViViT/L & \cmark \\  %
		& CLIP$\to$K700 & 1 &  $39.2$ & $44.5$ & 320 & ViViT/L & \cmark \\  %
		STAR/L (ours)$^{*}$ & CLIP$\to$K700 & 1 &  $\mathbf{41.7}$ & $44.5$ & 320 & ViViT/L & \cmark \\  %
		
		\bottomrule
	\end{tabular}
	} %
	\vspace{-0.5\baselineskip}
	\label{tab:sota_ava}
\end{table*}

Table~\ref{tab:sota_ava} shows that we achieve state-of-the-art results on both the challenging AVA and AVA-Kinetics datasets.
The previous best methods relied on external proposals~\cite{wang2022internvideo, wu2022memvit, arnab2022beyond} and external memory banks~\cite{pan2021actor, wu2022memvit} which we outperform.
There are fewer prior end-to-end approaches, and we outperform these by an even larger margin.
We achieve greater relative improvements on AVA-Kinetics, showing that our end-to-end approach can leverage larger datasets more effectively.
Note that we do not perform any test-time augmentation, in contrast to other approaches that ensemble results over multiple resolutions and/or left/right flips.
To our knowledge, we surpass the previous best reported results on these datasets, achieving a Frame AP of 41.7 on AVA, and 44.6 on AVA-Kinetics.
Notably, we outperform InternVideo~\cite{wang2022internvideo}, a recent video foundation model that is pretrained on 7 different web-scale video datasets.
The model of~\cite{wang2022internvideo} consists of two different encoders, one of which is also initialised from CLIP~\cite{radford2021learning}.
Like InternVideo, we achieve the best results on AVA by training a model on AVA-Kinetics, and then evaluating it only on the AVA validation set.

\vspace{\paravspace}
\paragraph{UCF101-24}

\begin{table*}[t]
	\centering
    \vspace{-0.2\baselineskip}
	\caption{Comparison to the state-of-the-art on datasets with tubelet annotations, namely UCF101-24~\cite{soomro_arxiv_2012} and JHMDB51-21~\cite{jhuang2013towards}.
	For Video~AP on UCF101-24, predicted tubelets of STAR models were linked using the causal algorithm from \cite{kalogeiton_iccv_2017,li2020actions} for fair comparison.
	For Video AP calculation on JHMDB, we processed the entire video with our network, and did not link tubelets.
	}
	\renewcommand{\arraystretch}{0.92}
	\begin{tabular}{llccccccccc}
		\toprule
		
		& & \multicolumn{4}{c}{UCF101-24} & \multicolumn{3}{c}{JHMDB51-21} \\
		\cmidrule(lr){3-6} \cmidrule(lr){7-9}
		
		& Pretraining & fAP & vAP20 & vAP50 & vAP50:95 & fAP & vAP20 & vAP50 & Backbone \\
		\midrule
		ACT~\cite{kalogeiton_iccv_2017} & IN1K & $67.1$ & $77.2$ & $51.4$ & $25.0$ & $65.7$ & $74.2$ & $73.7$ & VGG \\
		MOC~\cite{li2020actions} & IN1K $\to$ COCO & $78.0$ & $82.8$ & $53.8$ & $28.3$ & $70.8$ & $77.3$ & $77.2$ & DLA34 \\
		Unified~\cite{arnab2021unified} & K600 & $79.3$ &  -- & -- & -- & -- & -- & -- & SlowFast \\
		WOO~\cite{chen2021watch}  & K600 &  -- & -- & -- & -- &	80.5 & -- & -- & SlowFast \\
		TubeR~\cite{zhao2022tuber}      & IG65M$\to$K400             &  $83.2$  &  $83.3$  &  $58.4$  &  $28.9$  & --  & $87.4$ & $82.3$  &  CSN-152                 \\
		TubeR with flow~\cite{zhao2022tuber}           & K400     &  $81.3$  &  $85.3$  &  $60.2$     &  $29.7$  &  -- & $81.8$ & $80.7$  &  I3D   \vspace{0.3ex}    \\
		\midrule
		
		STAR/B (ours)  	 & IN21K$\to$K400  &  $87.3$  &  $87.7$  &  $66.2$  &  $30.9$  & $86.6$ & $89.1$ & $88.5$ & ViViT/B      \\  %
	     STAR/L (ours) & CLIP$\to$K700  &  $\mathbf{90.3}$  &  $\mathbf{88.0}$  &  $\mathbf{71.8}$  &  $\mathbf{35.2}$  & $\mathbf{92.1}$ & $\mathbf{93.1}$ & $\mathbf{92.6}$ & ViViT/L      \\  %
		\bottomrule
	\end{tabular}
	\vspace{-1\baselineskip}
	\label{tab:sota_ucf101_jhmdb}
\end{table*}

Table~\ref{tab:sota_ucf101_jhmdb} shows that we achieve state-of-the-art results on UCF101-24, both in terms of frame-level detection metrics (Frame AP), and tube-level detection metrics (Video AP).
We achieve state-of-the-art results when using a model pretrainined on Kinetics 400, and then see further improvements when using a larger backbone and pretraining on Kinetics 700, with original CLIP initialisation, consistent with our results on AVA (Tab.~\ref{tab:ablation_pretraining} and~\ref{tab:sota_ava}).
In particular, to our knowledge, we outperform the best previous reported Video AP50 number by 11.6 points.
Note that as UCF videos are a maximum of 900 frames, and our network processes $T = 32$ frames, we link together tubelets using the same causal algorithm as~\cite{kalogeiton_iccv_2017, li2020actions}.

\vspace{\paravspace}
\paragraph{JHMDB51-21}

Table~\ref{tab:sota_ucf101_jhmdb} also shows that we surpass the state-of-the-art on JHMDB, on both Frame AP and Video AP metrics too.
The videos in this dataset are trimmed (meaning that labelled actions are being performed on each frame), and also shorter.
As a result, the Video AP is not as strict as it is on UCF101-24.
Additionally, as the input videos are a maximum of 40 frames, we set $T = 40$ in our model so that we process the entire clip at once and therefore do not need to perform any tubelet linking.

\begin{figure}[t]
	\vspace{0\baselineskip}
	\includegraphics[width=\linewidth, trim={0 10.6cm 18.3cm 0}, clip]{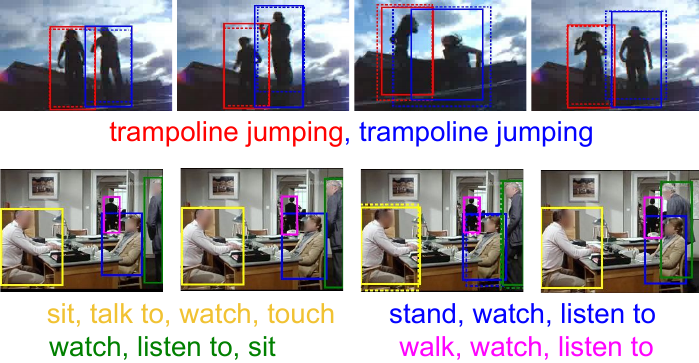}
	\vspace{-1\baselineskip}
	\caption{
        Visualisation of the tubelets predicted by our model.
        The colour corresponds directly to the spatial index of a query token, as our model produces tubelets without any postprocessing.
        Dashed lines denote the ground truth bounding box, and the predicted class label is below the tubelet.
        The first row shows a tubelet over 5.4 seconds on UCF101-24.
        Second row shows a tubelet over 2.6 seconds on AVA where only the central keyframe is annotated.
	}
	\label{fig:examples}
\end{figure}

\vspace{\paravspace}
\paragraph{Qualitative examples}
Figure~\ref{fig:examples} presents visualisations of our model's tubelets.%

\section{Conclusion}
\vspace{-0.5\baselineskip}

We have presented STAR, an end-to-end spatio-temporal action localisation model that can output tubelets, when either sparse keyframe, or full tubelet annotation is available.
Our approach achieves state-of-the-art results on four action localisation datasets for both frame-level and tubelet-level predictions (in particular, we obtain 44.6\% mAP on the challenging AVA-Kinetics dataset), outperforming complex methods that use external proposals and memory banks.

{\small
\bibliographystyle{ieee_fullname}
\bibliography{bibliography}
}

\appendix

\section{Additional experiments}
\label{sec:app_experiments}

\subsection{Implementation details}

We exhaustively list hyperparameter choices for the models used in our state-of-the-art comparisons in Tab.~\ref{tab:architecture-details} and~\ref{tab:hyperparamters}.

Note that our model hyperparameters in Tab.~\ref{tab:architecture-details} follow the same nomenclature from ViT~\cite{dosovitskiy_iclr_2021} and ViViT~\cite{arnab2021vivit} for defining ``Base'' and ``Large'' variants.

Our experiments use similar data pre-processing and augmentations as prior work~\cite{feichtenhofer_iccv_2019, wu2022memvit, wu_cvpr_2019}, such as horizontal flipping, colour jittering (consistently across all frames of the video) and box jittering.
In addition, we used a novel keyframe ``decentering'' augmentation (Sec.~\ref{sec:keyframe_decentering}) as our model predicts tubelets, and more aggressive scale augmentation (Sec.~\ref{sec:scale_aug}).

We train with synchronous SGD and a cosine learning rate decay schedule.
As shown in Tab.~\ref{tab:hyperparamters}, we typically use the same training hyperparameters across experiments.
Note that for the JHMDB dataset, we use $T = 40$ frames as input to our model, as this is sufficient to cover the longest video clips in this dataset.
We also do not need to perform ``decentering'' (Sec.~\ref{sec:keyframe_decentering}) for datasets with full tube annotations (UCF101-24 and JHMDB51-21).
As shown in Tab.~\ref{tab:architecture-details}, we found it beneficial to use a lower learning rate for the vision encoder of our model, as it was already pretrained, in contrast to the decoder which was learned from scratch.

\begin{table}[t]
\centering
\caption{Model architecture hyperparameters. 
We used the same decoder even when scaling up the vision encoder.
}
\begin{tabular}{lcc}
\toprule
\multirow{2}{*}{Hyperparameter} & \multicolumn{2}{c}{Model size} \\
\cmidrule(l){2-3}
               & Base         & Large         \\
\midrule

\textit{Decoder} \\
Number of layers &  \multicolumn{2}{c}{$6$} \\
Learning rate & \multicolumn{2}{c}{$10^{-4}$}  \\
Hidden size & \multicolumn{2}{c}{$256$} \\
MLP dimension & \multicolumn{2}{c}{$2048$} \\
Dropout rate & \multicolumn{2}{c}{$0.1$} \\
Box head num. layers & \multicolumn{2}{c}{$3$} \vspace{0.5ex} \\

\textit{Encoder} \\
Learning rate & $5\times10^{-6}$ & $2.5\times10^{-6}$ \\
Learning rate (CLIP init.) & \multicolumn{2}{c}{$1.25\times10^{-6}$} \\
Patch size &\multicolumn{2}{c}{$16\times16\times2$} \\
Spatial num. layers & $12$ & $24$ \\
Temporal num. layers & $4$ & $8$ \\
Attention heads & $12$ & $16$ \\
Hidden size & $768$ & $1024$ \\
MLP dimension & $3072$ & $4096$ \\
\bottomrule
\end{tabular}
\label{tab:architecture-details}
\end{table}

\begin{table*}[t]
\centering
\caption{Model training hyperparameters for the four datasets considered in our paper. We train with synchronous SGD and a cosine learning rate decay schedule.
}
\begin{tabular}{lcccc}
\toprule
\multirow{2}{*}{Hyperparameter} & \multicolumn{4}{c}{Dataset} \\
\cmidrule(l){2-5}
               & AVA & AVA-K & UCF101-24 & JHMDB51-21 \\
\midrule
Epochs (training steps) & $30$ ($148\,050$) & $30$ ($246\,690$) & $30$ ($88\,230$) & $40$ ($6\,800$) \\
Batch size & \multicolumn{4}{c}{$128$} \\
Optimiser & \multicolumn{4}{c}{Adam \cite{kingma2014adam}} \\
Adam $\beta_1$ & \multicolumn{4}{c}{$0.9$} \\
Adam $\beta_2$ & \multicolumn{4}{c}{$0.999$} \\
Gradient clipping $\ell_2$ norm & \multicolumn{4}{c}{1.0} \\
Focal loss $\alpha$  & \multicolumn{4}{c}{$0.3$} \\
Focal loss $\gamma$ & \multicolumn{4}{c}{$2.0$} \\
Number of spatial queries ($S$) &\multicolumn{4}{c}{$64$} \\
Number of frames ($T$) & $32$ & $32$ & $32$ & $40$ \\
Center deviation, $\rho$: per-frame matching & $4$ & $4$ & $0$ & $0$ \\
Center deviation, $\rho$: tubelet matching & $16$ & $16$ & $0$ & $0$ \\
Stochastic depth~\cite{huang_stochasticdepth_eccv_2016} & $0.2$ & $0.2$ & $0.5$ & $0.5$ \\

\bottomrule
\end{tabular}
\label{tab:hyperparamters}
\end{table*}

\subsection{Decentering}
\label{sec:keyframe_decentering}
\begin{table}[t]
\centering
\caption{Effect of keyframe decentering studied on the AVA dataset (resolution 160p) for a model with IN21K$\to$K400 initialisation, and factorised queries.
Mild amounts of keyframe decentering do not hurt performance measured on the center frame while explicitly supervising the models ability to localise and predict actions on other frames.
In fact, models trained with small amounts of decentering tend to perform better than models trained without any decentering.
}
\begin{tabular}{cc}
\toprule
Center deviation, $\rho$ & mAP~$\uparrow$ \\  %
\midrule
$0$         &  $26.5$  \\
$1$      &  $26.8$  \\
$2$      &  $26.5$  \\
$4$      &  $26.7$  \\
$8$      &  $26.4$  \\
$16$     &  $26.6$  \\
\bottomrule
\end{tabular}
\label{tab:ablation_decentering}
\end{table}

The majority of prior work on keyframe-based action localisation datasets (\eg AVA and AVA-Kinetics) predict only at the centre frame of the video clip, as only sparse supervision at this central keyframe is available.
As our model predicts \textit{tubelets}, we intuitively would like to supervise it for other frames in the input clip as well.

To this end, we introduce another data augmentation strategy, named ``decentering'', where we sample video clips during training such that the keyframe with supervision is no longer at the central frame, but may deviate randomly from the central position.
We parameterise this by an integer, $\rho$, which defines the maximum possible deviation, and randomly sample a displacement $\in [-\rho, \rho]$ during training.

We found that this data augmentation strategy results in qualitative improvements in the predicted tubelets (as shown in the supplementary video).
However, as shown in Tab.~\ref{tab:ablation_decentering}, it has minimal effect on the Frame AP which only measures performance on the annotated, central frame of AVA video clips.

Note that for datasets with full tube annotations, \ie UCF101-24 and JHMDB51-21, there is no need to apply decentering, as each frame of the video clip is already annotated.
We do, however, use decentering with the $\rho = 8$, when training with weak supervision on UCF101-24 (Tab.~4 of the main paper).

\subsection{Scale augmentation}
\label{sec:scale_aug}
\begin{table}[t]
\centering
\caption{Comparison of spatial scale augmentation for our models with a ViViT/B backbone on the AVA dataset (resolution of 140p on the shorter side).
We find that large range in scale-jittering, in the range of $(0.5, 2.0)$ of the original input frame, as used in~\cite{ghiasi2021simple} works the best.
Notably, doing scale augmentation in range of $(\frac{8}{7}, \frac{10}{7})$ as in the open-sourced SlowFast~\cite{feichtenhofer_iccv_2019} performs significantly worse.
Performing no scale augmentation (first row) performs the worst as expected.
}
\begin{tabular}{lc}
\toprule
Scale (min, max) & mAP~$\uparrow$ \\  %
\midrule
$(1, 1)$ (none) & 22.5 \\  %
$(1.14, 1.43)$~\cite{feichtenhofer_iccv_2019} & 23.9 \\  %
\midrule
$(0.25, 1.0)$         &  22.7  \\
$(0.5, 1.0)$          &  23.4  \\
$(0.25, 4.0)$         &  25.1  \\
$(0.5, 2.0)$          &  \textbf{25.6}  \\
\bottomrule
\end{tabular}
\label{tab:ablation_scale_aug}
\end{table}

Consistent with object detection in images~\cite{ghiasi2021simple,lin2017focal,cubuk_arxiv_2019,singh2018sniper}, we found it necessary to perform spatial scale augmentation during training to achieve competitive action localisation performance.
As shown in Tab.~\ref{tab:ablation_scale_aug}, we found that performing ``zoom out'' as well as ``zoom in'' scale augmentation during training significantly boosts action localisation performance.
This departs from the choice of performing ``zoom in'' only scale augmentation in previous work~\cite{feichtenhofer_iccv_2019, wu2022memvit, wu_cvpr_2019}.

\subsection{Focal and auxiliary loss}
\begin{table}[t]
\centering
\caption{Effect of using sigmoid loss and auxiliary losses studied on the AVA dataset (resolution 160p) for a model with IN21K$\to$K400 initialisation. Focal loss ($\alpha=0.3$ and $\gamma=2$) clearly performs better than the alternatives. Moreover, the use of auxiliary losses leads to a mild degradation in performance when combined with focal loss, but improves results when the focal loss is not used.}
\begin{tabular}{ccc}
\toprule
Focal loss & Auxiliary losses & mAP~$\uparrow$ \\  %
\midrule
\xmark & \xmark  &  $20.8$  \\
\xmark & \cmark  &  $21.8$  \\
\cmark & \cmark  &  $26.4$  \\
\cmark & \xmark  &  $\mathbf{26.8}$  \\
\bottomrule
\end{tabular}
\label{tab:ablation_aux_loss}
\end{table}

Following \cite{minderer2022simple, zhu2020deformable, zhang2022dino} we use sigmoid focal cross-entropy loss~\cite{lin2017focal} as our classification loss,
\begin{align}
\mathcal{L}_{\text{class}}(a, \hat{a})=&-\alpha \cdot a  \cdot \hat{a}^\gamma\log(\hat{a}) \\
                                   &-(1-\alpha)(1 - a)(1 - \hat{a})^\gamma\log(1 - \hat{a})\text{,} \nonumber
\end{align}
where $a$ and $\hat{a}$ are the ground truth and predicted action class probabilities respectively.
$\alpha$ and $\gamma$ are hyperparameters of the focal loss~\cite{lin2017focal}.
Furthermore, following \cite{minderer2022simple} we do not use auxiliary losses~\cite{carion_eccv_2020} (\ie attaching output heads after each decoder layer and summing up the losses from each layer) previously found to be beneficial for matching-based detection models.
Both of these choices are motivated by our ablations in Tab.~\ref{tab:ablation_aux_loss}:
We observe that the focal loss consistently improves performance, and that auxilliary losses are only beneficial when the focal loss is not used.

\end{document}